\documentclass[10pt,twocolumn,letterpaper]{article}

\usepackage{cvpr}
\usepackage{times}
\usepackage{epsfig}
\usepackage{graphicx}
\usepackage{amsmath}
\usepackage{amssymb}

\usepackage{graphicx}
\usepackage{latexsym}
\usepackage{amsmath}
\usepackage{amssymb}
\usepackage{xcolor}
\usepackage{paralist}
\usepackage{multirow}
\usepackage{threeparttable}
\usepackage[mathscr]{eucal}
\usepackage[ruled]{algorithm2e}
\newcommand*\samethanks[1][\value{footnote}]{\footnotemark[#1]}

\usepackage[breaklinks=true,bookmarks=false]{hyperref}

\cvprfinalcopy % *** Uncomment this line for the final submission

\def\cvprPaperID{6679} % *** Enter the CVPR Paper ID here

\ifcvprfinal\fi

\begin{document}

% \thispagestyle{empty}

% ******************************** Title ************************************
\title{CPR-GCN: Conditional Partial-Residual Graph Convolutional Network \\ 
in Automated Anatomical Labeling of Coronary Arteries}

%indicates equal contribution. This work is done when Xingjian Zhen was an intern at Alibaba Group
% \author{Han Yang\textsuperscript{$\flat,$}\protect \thanks{Equal Contribution.} ,
% Xingjian Zhen\textsuperscript{$\S,$}\samethanks[1] ,
% Ying Chi\textsuperscript{$\flat,$}\protect \thanks{Corresponding Author.} ,
% Lei Zhang\textsuperscript{$\flat$},
% and Xian-Sheng Hua\textsuperscript{$\flat$}\\
% \textsuperscript{$\flat$}Alibaba Group, Hangzhou, China, \textsuperscript{$\S$}University of Wisconsin Madison\\
% {\tt\small \{xiuxian.yh,xinyi.cy,lei.zhang.lz,xiansheng.hxs\}@alibaba-inc.com, \{xzhen3\}@wisc.edu}
% }
% \protect \thanks{Corresponding Author.} 
\author{Han Yang\textsuperscript{$1,$}\protect \thanks{Equal Contribution.} , 
Xingjian Zhen\textsuperscript{$2,$}\samethanks[1] , 
Ying Chi\textsuperscript{$1,$}\protect \thanks{Corresponding Author.} , 
Lei Zhang\textsuperscript{$1$}, 
and Xian-Sheng Hua\textsuperscript{$1$}\\
\textsuperscript{$1$}DAMO Academy, Alibaba Group ~~~ \textsuperscript{$2$}University of Wisconsin Madison\\
{\tt\small \{xiuxian.yh,xinyi.cy,lei.zhang.lz,xiansheng.hxs\}@alibaba-inc.com, xzhen3@wisc.edu}
}

\maketitle
\thispagestyle{empty}

% ***************************** Abstract ************************************
\begin{abstract}
Automated anatomical labeling plays a vital role in coronary artery disease diagnosing procedure. The main challenge in this problem is the large individual variability inherited in human anatomy. Existing methods usually rely on the position information and the prior knowledge of the topology of the coronary artery tree, which may lead to unsatisfactory performance when the main branches are confusing. Motivated by the wide application of the graph neural network in structured data, in this paper, we propose a conditional partial-residual graph convolutional network (CPR-GCN), which takes both position and CT image into consideration, since CT image contains abundant information such as branch size and spanning direction. Two majority parts, a partial-residual GCN and a conditions extractor, are included in CPR-GCN. The conditions extractor is a hybrid model containing the 3D CNN and the LSTM, which can extract 3D spatial image features along the branches. On the technical side, the partial-residual GCN takes the position features of the branches, with the 3D spatial image features as conditions, to predict the label for each branches. While on the mathematical side, our approach twists the partial differential equation (PDE) into the graph modeling. A dataset with 511 subjects is collected from the clinic and annotated by two experts with a two-phase annotation process. According to the five-fold cross-validation, our CPR-GCN yields 95.8\% meanRecall, 95.4\% meanPrecision and 0.955 meanF1, which outperforms state-of-the-art approaches.
\end{abstract}

% ***************************** Introduction ************************************
\section{Introduction}
Cardiovascular disease is one of the leading causes of death worldwide \cite{who2008}. Cardiac CT angiography (CCTA) image is widely adopted for the diagnosis of the cardiovascular disease because of its non-invasion and high sensitivity \cite{leipsic2014scct}. In the clinic, doctors need a series of manual labor to obtain the diagnostic report, which is a time-consuming effort. If a computer-aided diagnosis (CAD) system can generate the diagnostic report automatically, a huge amount of time can be saved. While automated anatomical labeling of the coronary artery tree extracted from CCTA image is a prerequisite step in the automated CAD system. 

\begin{figure}[t]
\setlength{\abovecaptionskip}{-0.1cm}
\setlength{\belowcaptionskip}{-0.5cm}
\centering
\includegraphics[width=0.9\columnwidth]{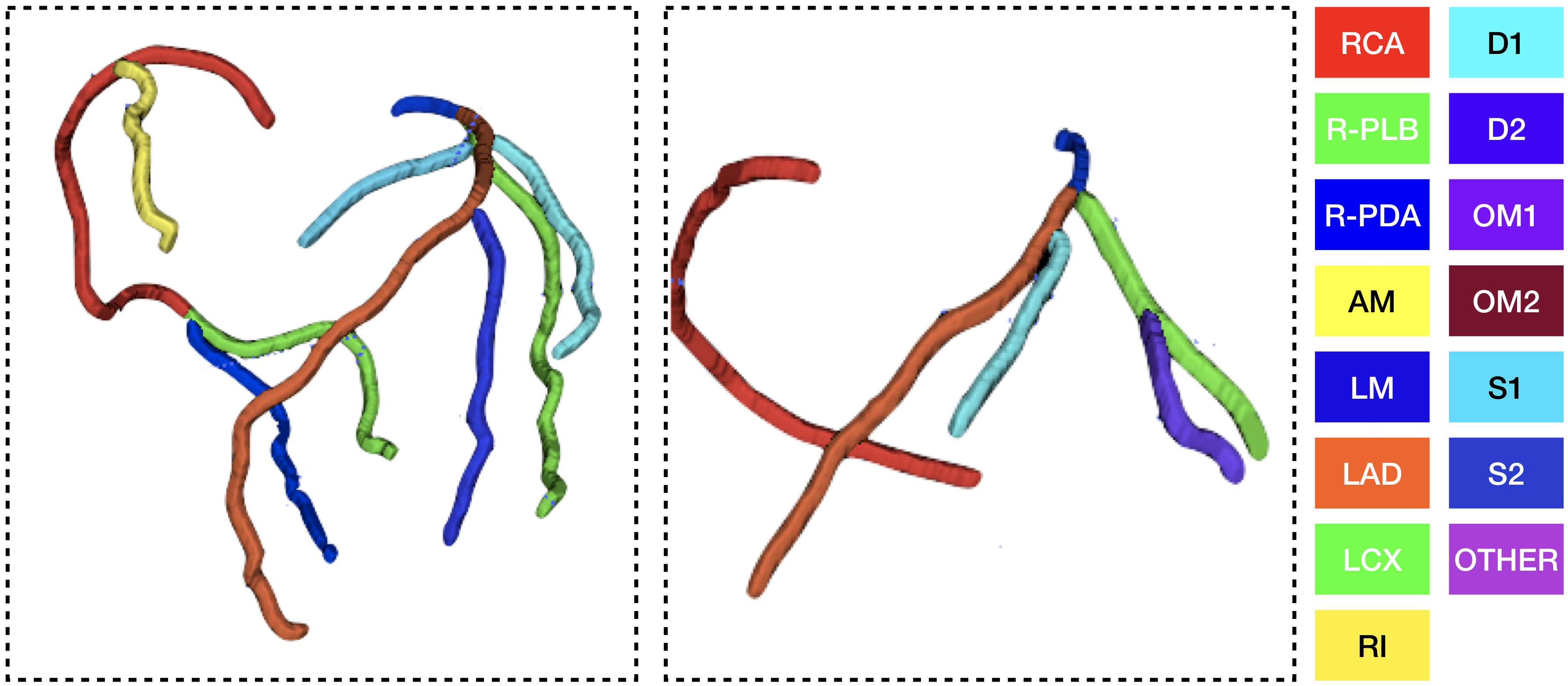}
\caption{Examples of coronary arteries for different subjects. Each color indicates a specific branch. The number of branches, the direction, and the connection all vary between these two subjects. The left vessel tree containing 10 branches is more complete, while multiple vessel branches are missing in the right one.}
\label{fig:intro_example}
\end{figure}

The coronary artery tree consists of two components, i.e., left domain (LD) and right domain (RD). Also they both originate from the aorta. According to \cite{yang2011automatic}, the main coronary arteries of interest are left main (LM), left descending artery (LAD), left circumﬂex artery (LCX), left ramus-intermedius (RI), obtuse margin (OM), diagonal artery (D), septal artery (S), right coronary artery (RCA), right posterior lateral branches (R-PLB), right posterior descending artery (R-PDA), right acute marginal artery (AM) (shown in Figure \ref{fig:intro_example}). From prior knowledge, we know that LAD, LCX, and RI are from LM. R-PLB, R-PDA, and AM are from RCA. Also S's and D's are from LAD, while OM's are from LCX. This makes the whole coronary artery tree as structured data. Normally, RCA, LM, LAD, and LCX are treated as the main branches. Other branches are treated as the side branches.

Several anatomical labeling techniques have been developed for coronary arteries \cite{yang2011automatic,cao2017automatic,wu2019automated}, brain arteries \cite{bogunovic2013anatomical}, abdominal arteries \cite{matsuzaki2015automated,zhang2013automatic} and airways \cite{gu2012automated}. However, as shown in Figure \ref{fig:intro_example}, the coronary arteries vary much among subjects, which is the main challenge for the labeling system. The number of branches, the length and size of each branch, and the direction that the branch span all vary from person to person. \cite{yang2011automatic,cao2017automatic} both rely on registration algorithm and prior knowledge. They first identified four main branches (i.e, LM, LAD, LCX and RCA) and then labeled the side branches (e.g., AM, R-PDA, D, etc.). Finally, the results were refined by logical rules which are translated from the clinical experience. But these conventional methods are not data-driven, i.e., they can not leverage the advantage of big data. Recently, in \cite{wu2019automated}, the author brought a novel deep neural network (TreeLab-Net), which can learn from the position features extracted from the coronary artery centerlines for labeling the segments. Although, this model is data-driven, it only utilizes position information from vessel centerline and leaves the full information in the CCTA images aside. In addition, the input for the TreeLab-Net is built by topological structure. Missing the main branches (e.g., LM, RCA, etc.) might have deep influence in labeling the side branches.  Therefore, a robust and self-adaptive model is needed for this structured data.

In the deep learning field, it's quite mature for the study in Euclidean space where elements are treated equally. However, there are normally two view aspects for the structured data. In \cite{chakraborty2018statistical,chakraborty2019deep}, the authors viewed the structured data from a manifold-valued aspect. Also in \cite{kipf2016semi,henaff2015deep,zhou2018graph} where the structured data can be viewed as the graph, the authors introduced the graph models with nodes and edges, which can be used to extract the information from the relationship between each node in the structured graph data.
It's natural to treat the coronary arteries as the tree because of the path the branches spread and the connection among the branches. 

In this paper, we propose a conditional partial-residual graph convolutional network (CPR-GCN), which can make full use of both position information and 3D image information in the CCTA volume. The partial-residual block is applied to the position domain features to enhance the features. Also, we use 3D Convolutional Neural Network (CNN) together with Bidirectional Long Short-Term Memory (BiLSTM) to extract the features along each branches as the conditions for the graph model. These two parts compose the CPR-GCN which can be trained end-to-end. 

In a summary, our main contributions are as follows:
\begin{itemize}
\item We propose the CPR-GCN, a conditional partial-residual graph convolutional network, which can label the coronary artery tree end-to-end.
\item To our best knowledge, this is the first time we take 3D image features into consideration in coronary artery labeling field. 
\item Our CPR-GCN and the hybrid model (i.e., 3D CNNs following BiLSTMs) can be jointly trained. We evaluate the CPR-GCN on a large private collected dataset. Our approach outperforms the state-of-the-art result.
\end{itemize}

% In a summary, our main contributions are as follows:
% \begin{itemize}
% \item For this specific Coronary Arteries labeling problem, we introduce the extra information from 3D CCTA images, while almost all the other models only use the centerlines.
% \item Our CPR-GCN and the hybrid model can be trained and tested end-to-end.
% \item Our approach outperforms the SOTA methods, and is more robust when some vessels are missing.
% \end{itemize}

% ***************************** Related Work ************************************
\section{Related Work}
Most of the related work can be divided into two categories, i.e., traditional based and deep learning based. Traditional methods are based on the knowledge and the topology of the coronary arteries. Normally, they require two steps of registration and correction. With the development of deep learning, there are also some methods on the graph-based structural data. These methods extract features as nodes and train the model with the data they acquire. Also these methods heavily depend on the size and quality of the dataset.

\subsection{Traditional Methods}
Most traditional methods are based on registration. 
In \cite{yang2011automatic}, the author presented a two-step method. In the registration step, the main branches, LM, LAD, LCX, and RCA, are identified. Then the rest branches are matched afterward. 
In \cite{cao2017automatic}, they built the 3D models for both right dominant and left dominant. The 3D coronary trees from subjects are aligned with the 3D models to get the label of each segment. They also applied the logical rules to fulfill the clinical experience.

\begin{figure*}[!t]
\setlength{\abovecaptionskip}{-0.1cm}
\setlength{\belowcaptionskip}{-0.5cm}
\centering
\includegraphics[width=0.9\textwidth]{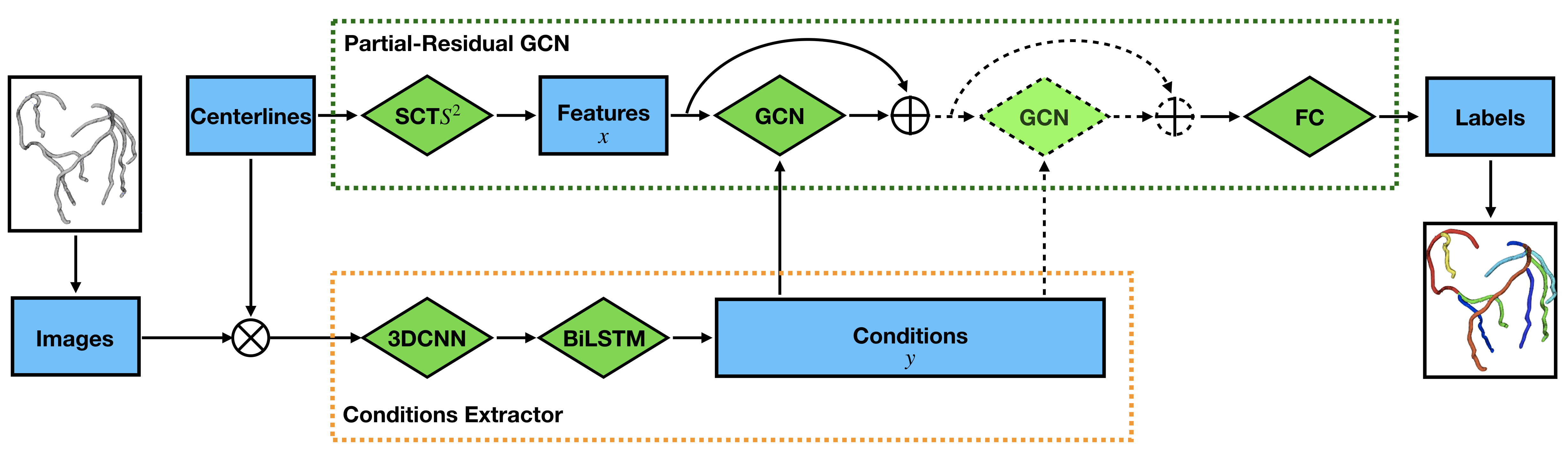}
\caption{The framework for the CPR-GCN. The two parts, conditions extractor and partial-residual GCN, compose our CPR-GCN model. The backbone of our model is in the green bounding box. The orange bounding box extracts extra information from image domain. The $\oplus$ is the residual connection block on the features $x$. The $\otimes$ means using control points along the centerlines to extract the moving cubes from CCTA images. }
\label{fig:method_framework}
\end{figure*}

However, traditional methods highly rely on the main branches. If the main branches are missing in the automated segmentation system, the performance will deteriorate dramatically. Also, they require prior knowledge about how the coronary tree span. Thus if some sub-branches are missing, it will affect the topology and the performance. Finally, all these traditional methods have the human-interpretation, which means that if the topology is too complicated, the automatic system will raise the information that it is not capable of determining. 

\subsection{Deep Learning Based Methods}
With the power of deep learning, the TreeLab-Net is developed in \cite{wu2019automated}. They combined the multi-layer perception encoder network and the bidirectional tree-structural LSTM to construct the TreeLab-Net. They used several selected features from positions and stacked the Child-Sum Tree LSTMs as the components. The left and right coronary arteries are trained independently.

In this method, the missing branches will cause a massive problem since this will change the layer index of the nodes inside the tree. With the tree-structured model, the message is only passing between nearby layers. The closer the branches are to the root nodes, the higher impact they will raise. The left and right are also classified with the rules and the thresholds. This method has a strong assumption that the branches will only bifurcate so that each node in the tree-structure can only have two children. But in the coronary arteries, it's quite reasonable that several sub-branches bifurcate from the points near to each other. So the node for the parent branch will return to have more than two children, which is beyond the capacity of the TreeLab-Net. 

In a broad sense, the TreeLab-Net is a simplified version of the graph models. In \cite{kipf2016semi}, the author brought out the Graph Convolutional Networks (GCN) which operate directly on graphs. By projecting the graphs into the Fourier domain, the author defined the convolution operator and filter kernels in the Fourier domain using the Chebyshev polynomials. In \cite{li2015gated}, they modified the Graph Neural Networks (GNN) \cite{scarselli2008graph} to use gated recurrent units. 

Even though these graph models are successful in molecular fingerprints \cite{duvenaud2015convolutional} and protein interface prediction \cite{fout2017protein}, they have not been used in labeling coronary arteries. Also graph models also suffer from shallow structure problems. Since stacking multiple GCN layers will result in over-smoothing \cite{li2018deeper}.

% ***************************** our Approach ************************************
\section{Our Approach}
In this section, we detail the CPR-GCN, which makes full use of both CCTA images and the position of the coronary artery centerlines. As shown in Figure \ref{fig:method_framework}, our method considers both CCTA images and the features from positions mentioned in \cite{wu2019automated}.
The centerlines are extracted using the automated coronary artery tracking system \cite{yang2019discriminative}. Then, our CPR-GCN extracts the features from the centerlines within SCT$S^2$ block. Also, in the image domain, we use the sub-sampled control points on centerlines to get the moving cubes with a fixed radius $\gamma$ along each branches as the image domain data. The conditions for our CPR-GCN model are obtained with the 3D CNN and the BiLSTM. The detailed architecture of the CPR-GCN model is shown in Table \ref{detail_model}.

\begin{table}[t]
% \footnotesize
\caption{The details of the parameters in our model. The framework is shown in Figure \ref{fig:method_framework}.}\smallskip
\centering
\resizebox{.95\columnwidth}{!}{
\smallskip\begin{tabular}{r | l}
\hline
              Block         & Details\\
\hline
  \multirow{2}{*}{SCT$S^2$} & first, middle, and last points\\
                            & tangent direction and first-last direction\\
\hline
\multirow{6}{*}{3D CNN}    & kernel size = 3, in channel = 1, out channel = 16\\
                            & maxpooling size = 2\\
                            & kernel size = 3, in channel = 16, out channel = 32\\
                            & maxpooling size = 2\\
                            & kernel size = 3, in channel = 32, out channel = 64\\
                            & maxpooling size = 2\\
\hline
                BiLSTM      & layer = 4, hidden size = 128\\
\hline
\multirow{2}{*}{CPR-GCN}    & out channel = 256\\
                            & out channel = 256\\
                            & out channel = 256\\
\hline
\multirow{2}{*}{Fully Connected} & out channel = 128\\
                                 & out channel = \# of classes\\
\hline
\end{tabular}
}
\label{detail_model}
\end{table}

\subsection{Position Domain Features}
In \cite{wu2019automated}, the author introduced a spherical coordinate transform 2D (SCT2D) which transforms the positions $P_k=[(x_i,y_i,z_i)]_{i=1}^{{\rm Length}_k}$ in 3D into the azimuth and elevation angles $[(\varphi_i,\theta_i)]_{i=1}^{{\rm Length}_k}$. They argued that this could normalize the variance of centerlines in the default Cartesian coordinate system. However, it is noticeable that $\varphi$ and $\theta$ have the periodic $2\pi$. So with limiting the range of the angle to be $[0,2\pi)$, the small amount of shaking, due to the noise, will return to be $2\pi$ difference near the angle $0$. 

The similar idea of using spherical coordinate transform is applied in our approach. Since each branches are processed separately, so as to get the azimuth and elevation angles, we need to define the origin and the $x,y,z$ axes for each branches. For each branches, the first control point is chosen as the origin. The direction pointing from the first point to the second point is defined as $z$ axis. The vector from the first point to the last point of the centerline lies in $y-z$ plane. 

To overcome the instability due to the periodic, we use the $S^2$ manifold to represent $\varphi, \theta$. The $S^2$ manifold is the sphere with unit radius in $\mathbb{R}^3$. A trivial method is to use the $2\times 2$ matrix $M =\begin{bmatrix} 
\rm{sin}\theta & \rm{sin}\varphi \\
\rm{cos}\theta & \rm{cos}\varphi 
\end{bmatrix}$. Because of the periodic of ${\rm sin}(\cdot)$ and ${\rm cos}(\cdot)$, the matrix $M$ is stable on the whole manifold $S^2$. 
This kind of spherical coordinate transformation is called SCT$S^2$ in the rest of this paper.
The Cartesian coordinate and the $S^2$ manifold transformation is in Eq. \ref{xyz2s2}.

\begin{align}
\label{xyz2s2}
    x &= r \rm{sin}\theta \rm{cos}\varphi &&r=\sqrt{x^2+y^2+z^2} \nonumber\\
    y &= r \rm{sin}\theta \rm{sin}\varphi &&\rm{cos}\theta=z/r, \theta \in [0,\pi]\\
    z &= r \rm{cos}\theta                 &&\rm{sin}\varphi=x/(r\rm{sin}\theta),\rm{cos}\varphi=y/(r\rm{sin}\theta) \nonumber
\end{align}

We use the similar features mentioned in \cite{wu2019automated}: \begin{inparaenum}[(1)]\item The $S^2$ projection and the normalized 3D positions of the first point, center point, and the last point. \item Directional vector between first and last points and the tangential direction at the start point in both 3D and $S^2$.\end{inparaenum}

\subsection{Image Domain Conditions}
\begin{figure}[!t]
\setlength{\abovecaptionskip}{-0.1cm}
\setlength{\belowcaptionskip}{-0.5cm}
\centering
\includegraphics[width=0.9\columnwidth]{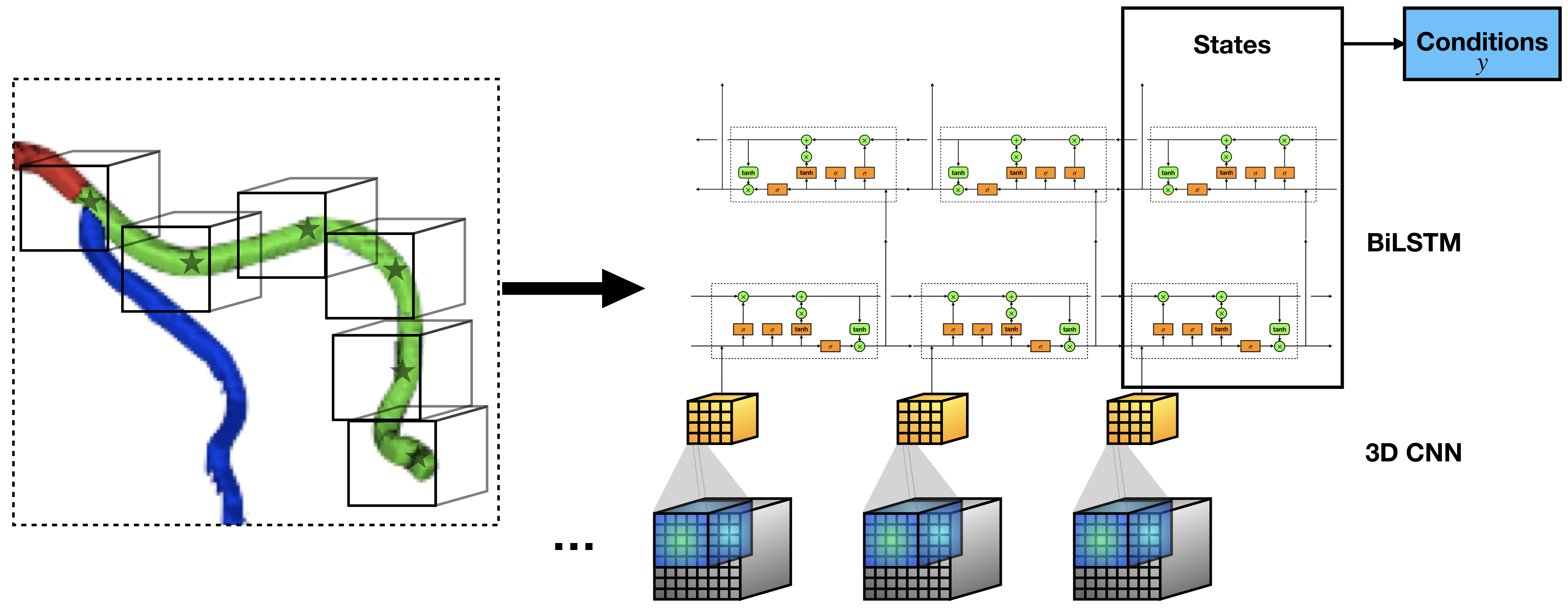}
\caption{The details of the framework in the image domain. We use 3D CNN and the BiLSTM to learn the conditions in the image domain for our partial-residual block. We use the last state as the final representative of the conditions.}
\label{fig:method_imageframework}
\end{figure}

Most of the medical images, including Magnetic Resonance Imaging (MRI) and the CCTA we use, lie in 3D. 
The branches, unlike other images, have the sequential dependency. 
Thus we use the 3D CNN to extract the spatial features and use the BiLSTM afterward to summarize the tubular sequential features. 
An example of processing the R-PLB branch is shown in Figure \ref{fig:method_imageframework}.
The $z$ dimension of CCTA images are the slices, which might have different spacing from $x$ and $y$ dimensions. So we resample all the CCTA images to have the same spacing $v$ along $x,y,z$ dimensions.

Using the automatic segmentation method, we can get the centerlines of all branches $(P_1,P_2,...,P_n)$ which build into a coronary artery tree. We separate the branches where the centerlines end or bifurcate. If the starting points of the children branches are close to each other, we consider all these children branches are from the same point on the parent branch. The control points of the centerlines are smoothed using the Catmull-Rom spline \cite{twigg2003catmull}. Finally, the control points are sub-sampled with the same length. 

The image domain data is the cubes $I_k$ with the fixed radius $\gamma$ around each control points of $P_k$. Three layers of 3D CNN and 3D maxpooling are used to extract the features of $I_k$. The weights of CNN are shared among the segments. 
In order to train the model in a mini-batch manner, these feature vectors are padded to the ${\rm max}_{k=1}^{n}({\rm Length}_k)$ as the input of the multi-layer bidirectional LSTM \cite{graves2005framewise}. 
The last hidden state is treated as the final conditions $y$, which represents the image information of this branch. We treat this as the conditional information since the images are of less importance than the position domain features. 

\subsection{Partial-Residual Block of GCN}
The layer-wise propagation rule for the traditional GCN is Eq. \ref{gcn}. In the multi-layer GCN, the features $X$ of nodes is the input for the first layer, $X \in \mathbb{R}^{n\times d_0}$. 
Here, the $n$ is the number of nodes. The $d_0$ is the dimension of the features for each nodes. The $A$ is the adjacency matrix for the graph. The $W^l$ is the layer-wise trainable weights, $W^l\in \mathbb{R}^{d_l\times d_{l+1}}$. The $\sigma(\cdot)$ is the activation function. In this paper, we choose ${\rm{ReLU}}(\cdot)={\rm max}(0,\cdot)$ as our activation function to include the nonlinear ability. 
\begin{align}
\label{gcn}
    &H^{(l+1)}=\sigma(\tilde{D}^{-\frac{1}{2}} \tilde{A} \tilde{D}^{-\frac{1}{2}} H^l W^l)\\
    &\tilde{A}=A+I_N ,
    \tilde{D}_{ii}=\sum_j \tilde{A}_{ij} \nonumber
\end{align}
The $\tilde{A}$ is the adjacency matrix $A$ added the self-loop identity matrix $I_N$. The input for the first layer is $H^0=X$.

\begin{figure}[t]
\setlength{\abovecaptionskip}{-0.1cm}
\setlength{\belowcaptionskip}{-0.5cm}
\centering
\includegraphics[width=0.6\columnwidth]{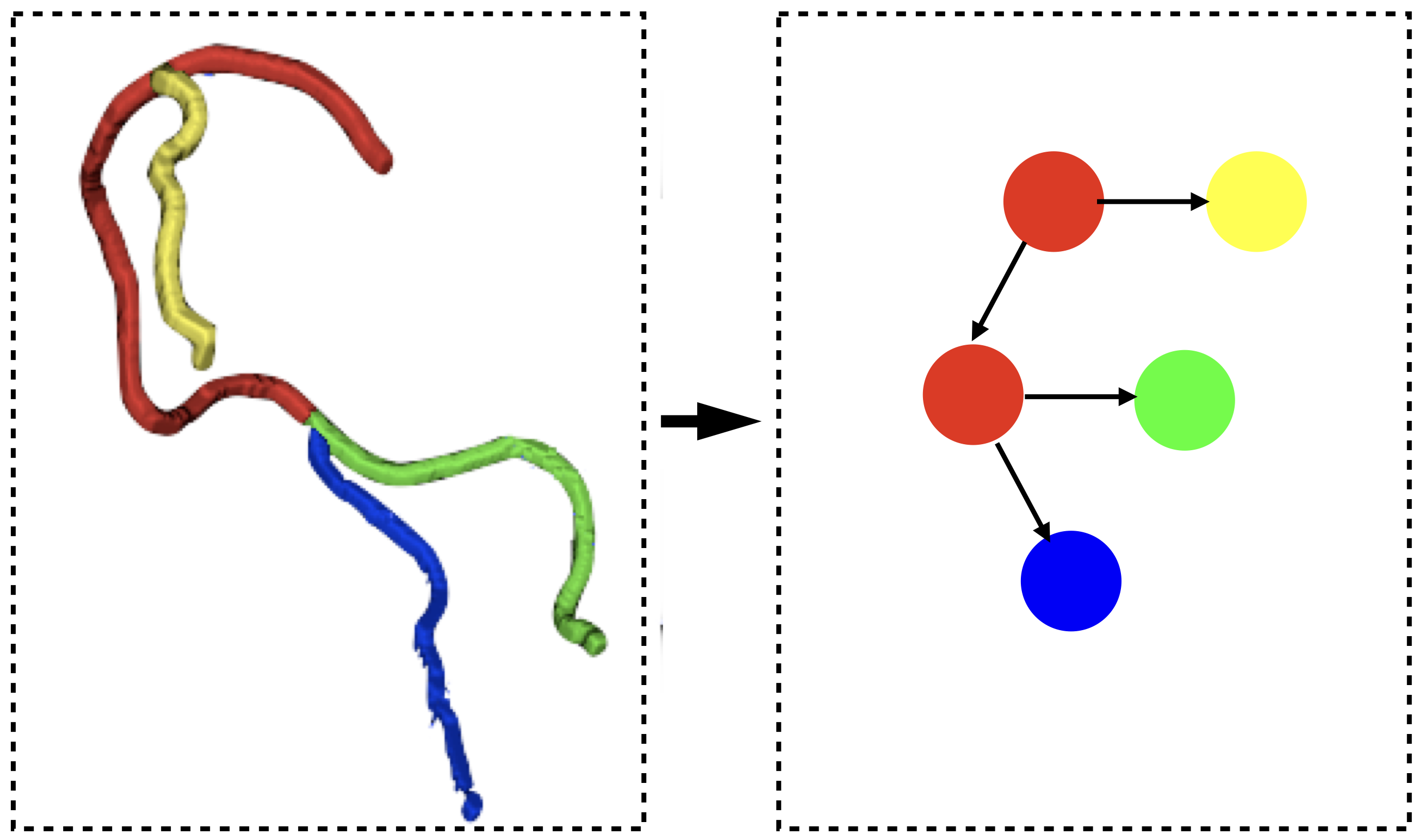}
\caption{The rules we used to build the graph for the CPR-GCN. Whenever the branches bifurcate, new node is added in the graph. So main branches (e.g., RCA, LAD, etc.) might be represented by multiple nodes in the graph. For example, two red nodes in the right box belong to RCA.}
\label{fig:method_graphframework}
\end{figure}

\begin{figure}[b]
\setlength{\abovecaptionskip}{-0.1cm}
\setlength{\belowcaptionskip}{-0.5cm}
\centering
\includegraphics[width=1.0\columnwidth]{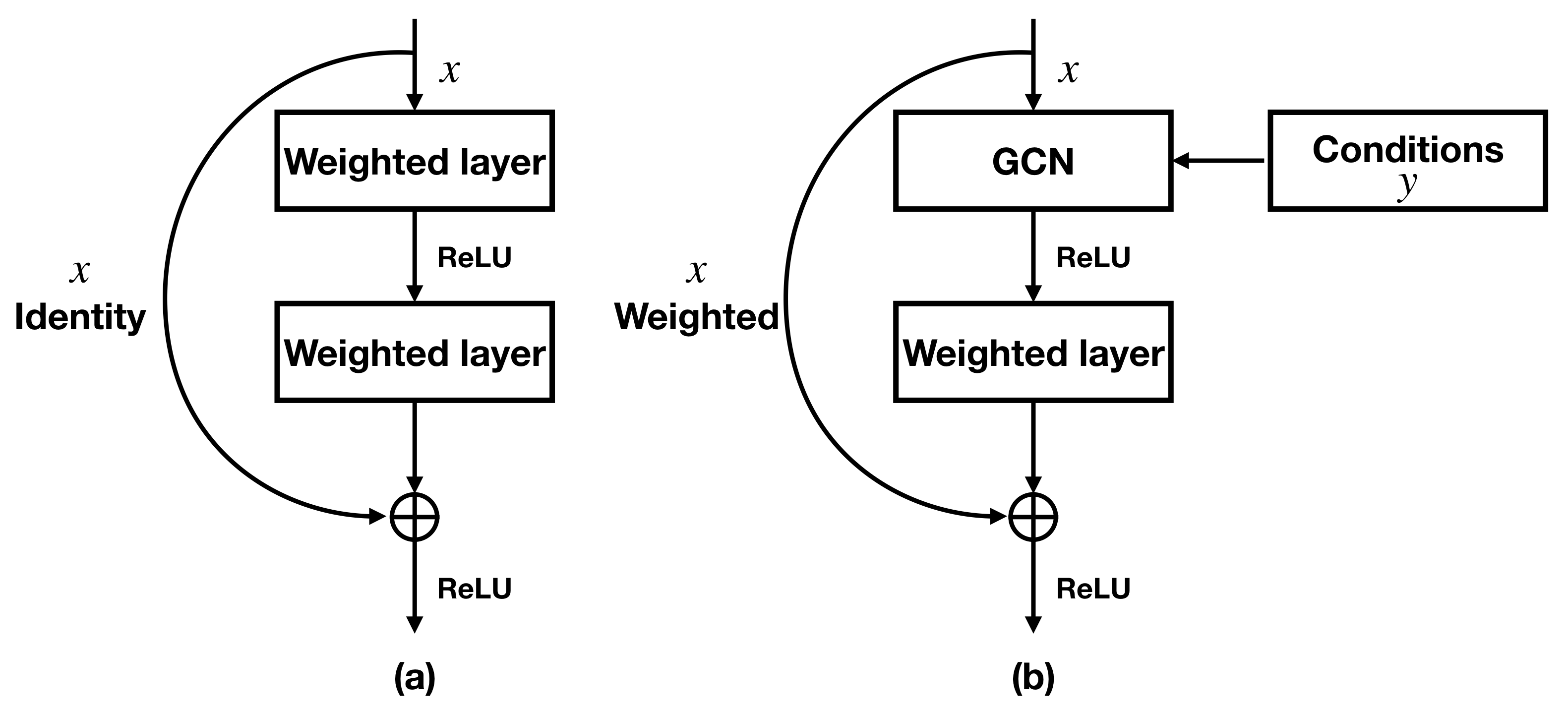}
\caption{The partial-residual block in the CPR-GCN. It can strengthen some part of the features to have more influence on the final layer while absorbing the other as the conditions. (a) is the traditional residual block, (b) is our partial-residual.}
\label{fig:method_residualblock}
\end{figure}

Our conditional partial-residual block requires both the position features $x$ and the CCTA image domain conditions $y$. The combination of the features and conditions from two domains is used as the representative of the nodes in the graph model. The edges are defined as the parent-children relationship. As we mentioned above, the topology of the whole coronary tree is collected from $(P_1,P_2,...,P_n)$. Whenever the branches bifurcate, we treat these parent-children branches as three (or more) nodes. The edges are from the parent branch to the children branches. This will build the graph of the subject with the adjacency matrix $A$. As shown in Figure \ref{fig:method_graphframework}, the RCA first bifurcates the AM and then bifurcates the R-PLB and R-PDA. So the extracted graph has 4 edges and 5 nodes. 

In \cite{he2016deep}, the author argued that the deep residual learning framework (Figure \ref{fig:method_residualblock} {\bf{(a)}}) can help in the performance. Instead of directly learning the map $\mathcal{H}(\mathbf{x})$, the neural network learns the residual part $\mathcal{F}(\mathbf{x}):=\mathcal{H}(\mathbf{x})-\mathbf{x}$ assuming the input and the output having the same dimension. When changing the output dimension, it's a straightforward method to add the linear projection $W_s$ by the shortcut connection:
\begin{align}
    {\mathbf{z}}=\mathcal{F}(\mathbf{x},{W_i})+W_s \mathbf{x}
\end{align}
Also in \cite{bresson2017residual}, the authors extended the idea of residual connection into the Residual Gated Graph ConvNets. The updated propagation rule is:
\begin{align}
\label{res_gcn}
    &H^{(l+1)}=\sigma(\tilde{D}^{-\frac{1}{2}} \tilde{A} \tilde{D}^{-\frac{1}{2}} H^l W^l) + H^l
\end{align}
If we view the residual connection in Eq. \ref{res_gcn} as a continuous function of $l$ and add enough number of layers. In the limit, \cite{chen2018neural} parameterized the continuous dynamics of hidden units using the ordinary differential equation (ODE):
\begin{align}
\label{ode_gcn}
    &\frac{dH(l)}{dl}=f_A(H(l),W(l),l)
\end{align}
In our setup, we have two kinds of input: position domain features $x(l)$ and the CCTA image domain conditions $y$. If we treat $l$ as the layer index number and $x$ is the function of layer $l$ in Eq. \ref{ode_gcn}, we have the partial differential equation (PDE) on $x,y$:
\begin{align}
    \label{pde_gcn1}
    \nabla H^l(x(l),y) &= \frac{\partial H^l(x,y)}{\partial x} dx + \frac{\partial H^l(x,y)}{\partial y} dy\\
    \label{pde_gcn2}
    &= \frac{\partial H^l(x,y)}{\partial x} \frac{dx}{dl}dl\\
    \label{pde_gcn3}
    & = {\rm{GCN}}_A(x,y)dl
\end{align}
Eq. \ref{pde_gcn2} is based on the fact that we treat $y$ as the conditions. 
We use the trainable ${\rm{GCN}}_A(x,y)$ to approach the partial differential $\frac{\partial H^l(x,y)}{\partial x} \frac{dx}{dl}$. 
If we take $H^0=x(0)=X$ and approximate $dl=1$, we have
\begin{align}
    H^1 &=\nabla H^l(x,y)dl+H^0 \nonumber\\
        &={\rm{GCN}}_A(x,y)+X
\end{align}
as the discrete numerical estimates. In our case, as shown in Figure \ref{fig:method_residualblock} {\bf{(b)}}, the discrete partial-residual block of the CPR-GCN takes the weighted $X$ because of the change of the channel sizes. 
If we push our PDE a bit further, we can have:
\begin{align}
    H^k &=\int \nabla H^l(x,y) dx + H^0 \nonumber\\
        &= \int_0^k {\rm{GCN}}_A(x,y)dl + X
\end{align}
Also, the $X$ here should be weighted for the flexibility of the channels. 

\subsection{Data Flow}
The algorithm of our CPR-GCN is shown in Alg. \ref{algorithm}. CCTA image $I$, centerlines $P_1, P_2, ..., P_n$ and reference labels $GT_1,GT_2,...,GT_n$ compose the training sample of the model. We first build the graph $A$ via the bifurcation of $P_1, P_2, ..., P_n$. The position domain feature $x$ and the image domain feature $y$ are extracted via SCT$S^2$ and a hybrid network (i.e., 3D CNN following BiLSTMs). In addition, we concatenate $x$ and $y$ as the input of GCN layers with $x$ as the shortcut in residual connection. At last, a fully connected layer predicts the final labeling $L_1,L_2,...,L_n$ and our object is to minimize the cross-entropy of these two distributions, i.e., $GT_1,GT_2,...,GT_n$ and $L_1,L_2,...,L_n$.

\begin{algorithm}[h]
\footnotesize
\caption{Training procedure of our approach}\label{algorithm}
\KwData{CCTA image ($I$), centerlines ($P_1, P_2, ..., P_n$)}
\textbf{Ground truth:} $GT_1,GT_2,...,GT_n$\\
$A\leftarrow$ Build\_Graph$(P_1, P_2, ..., P_n)$\;
$x_1, x_2,...,x_n \leftarrow$ SCT$S^2(P_1, P_2, ..., P_n)$\;
$I_1, I_2,...,I_n \leftarrow I \otimes (P_1, P_2, ..., P_n)$\;
$y_1,y_2,...,y_n\leftarrow$ BiLSTM $\leftarrow$ 3D CNN($I_1, I_2,...,I_n$)\;
$H\leftarrow$GCN$_A(x,y)+x$\;
$L_1,L_2,...,L_n\leftarrow$ FC$(H)$\;
loss$\leftarrow$Cross\_Entropy($L,GT$)\;
\end{algorithm}

%***************************** Experimental Results *************************************
\section{Experimental Results}

% \begin{table}[b]
% \caption{The basic information for the dataset we use. Segments are the branches after bifurcating. The edges are the relationship between segments.}\smallskip
% \centering
% \resizebox{.95\columnwidth}{!}{
% \smallskip\begin{tabular}{ccccc}
% \hline
% & Number & Avg.(std)) & Max  & Min\\
% \hline
% Branches     & 4929   & 9.65 (2.13)      & 15   & 3 \\
% Segments    & 6760   & 13.23 (3.55)     & 22   & 3 \\
% Edges       & 5714   & 11.18 (3.56)     & 20   & 2 \\
% \hline
% \end{tabular}
% }
% \label{table:dataset}
% \end{table}

\begin{table}[b]
% \footnotesize
\centering		
\caption{The basic information for the dataset we use. Segments are the branches after bifurcating. The edges are the relationship between segments.}\smallskip
\label{table:dataset}	
\smallskip
\resizebox{.8\columnwidth}{!}{
\begin{tabular}{ccccc}
\hline
& Number & Avg.(std)) & Max  & Min\\
\hline
Branches     & 4929   & 9.65 (2.13)      & 15   & 3 \\
Segments    & 6760   & 13.23 (3.55)     & 22   & 3 \\
Edges       & 5714   & 11.18 (3.56)     & 20   & 2 \\
\hline
\end{tabular}
}
\end{table}

\renewcommand{\arraystretch}{1.2}
\begin{table*}[t]
% \footnotesize
\caption[Caption for LOF]{Comparisons of conventional method \cite{cao2017automatic}, deep learning based TreeLab-Net \cite{wu2019automated} and our CPR-GCN on our dataset. Recall, precision and F1 score are used as the evaluation metrics.}
\centering
\label{table:main_compare}
\resizebox{1.0\textwidth}{!}{
\begin{tabular}{c|c|ccccccccccc|c}
\hline
Method & Metric & RCA &  R-PDA & R-PLB & AM & LM & LAD & LCX & RI & D & OM & S & Avg(std) \\
\hline %Cao et al. \cite{cao2017automatic}
\multirow{3}{2.5cm}{Conventional \cite{cao2017automatic}} & Recall & 0.918 & 0.850 & 0.852 & 0.893 & 0.984 & 0.911 & 0.832 & 0.848 & 0.799 & 0.720 & 0.835 & 0.859$\pm$0.066 \\
& Precision & 0.925 & 0.835 & 0.860 & 0.871 & 0.991 & 0.929 & 0.810 & 0.803 & 0.781 & 0.739 & 0.865 & 0.855$\pm$0.069 \\
& F1 & 0.922 & 0.842 & 0.856 & 0.882 & 0.987 & 0.920 & 0.821 & 0.825 & 0.789 & 0.730 & 0.850 & 0.857$\pm$0.067 \\
\hline %Wu et al. \cite{wu2019automated}
\multirow{3}{2.5cm}{TreeLab-Net \cite{wu2019automated}} & Recall & 0.950 & 0.858 & 0.818 & 0.871 & 0.996 & 0.948 & 0.913 & 0.770 & 0.816 & 0.805 & 0.862 & 0.873$\pm$0.067 \\
& Precision & 0.948 & 0.823 & 0.842 & 0.871 & 0.970 & 0.937 & 0.936 & 0.714 & 0.841 & 0.807 & 0.859 & 0.868$\pm$0.072 \\
& F1 & 0.949 & 0.840 & 0.830 & 0.871 & 0.983 & 0.942 & 0.924 & 0.741 & 0.829 & 0.807 & 0.860 & 0.871$\pm$0.069\\
\hline
\multirow{3}{2.5cm}{\textbf{Our CPR-GCN}} & Recall & 0.994 &0.930&	0.944&	0.991&	0.994&	0.990&	0.982&	0.921&	0.936&	0.896&	0.954& \textbf{0.958}$\pm$\textbf{0.033} \\
& Precision & 0.987&	0.946&	0.947&	0.983&	0.984&	0.986&	0.971&	0.896&	0.883&	0.933&	0.974 & \textbf{0.954}$\pm$\textbf{0.035} \\
& F1 & 0.990&	0.938&	0.945&	0.987&	0.989&	0.988&	0.976&	0.909&	0.909&	0.914&	0.964 & \textbf{0.955}$\pm$\textbf{0.032}\\
\hline
\end{tabular}
}
\end{table*}

\subsection{Dataset and Evaluation Metrics}
To our best knowledge, a public dataset with CCTA image and annotation of coronary artery labeling is not available until now. Previous works \cite{yang2011automatic,cao2017automatic,wu2019automated} all collected a private experimental dataset from clinic. For instance, conventional methods \cite{yang2011automatic,cao2017automatic} only used 58 and 83 subjects respectively, while deep learning based method \cite{wu2019automated} used a larger dataset with 436 subjects. In this study, we collected the largest relevant dataset from clinic. All vessel centerlines are first extracted using\cite{yang2019discriminative}. This dataset contains 511 subjects and all of them are annotated by two experts with a two-phase annotation process. These two experts give a label to every branch alone in the first round. Then the annotation results are merged and experts take discussions on inconsistent ones to obtain a final label. These 511 subjects and corresponding annotation compose our experimental dataset. The average number of branches in each subject is 9.65, with standard deviation (std) 2.13. The largest number of branches is 15, and the smallest number is 3. After bifurcated and separated, the average number of segments is 13.23. The detail is mentioned in Table \ref{table:dataset}. The edges in the table represent the relationship in the graph we build for each subject. The average number of edges in each graph is 11.18.

The evaluation is performed on all branch segments by the predicted label and the ground truth label. The recall rate for each segments is calculated by ${\rm Recall}=\frac{t_p}{t_p+f_p}| ({{\rm label}=i})$.
The precision is ${\rm Precision} =\frac{t_p}{t_p+f_n}| ({{\rm label}=i})$.
The F1 score is ${\rm F1} = 2\frac{\rm Precision \times Recall}{\rm Precision + Recall}|({\rm label}=i)$. Since the numbers of segments in are imbalance, we also use the mean metrics of all classes for the segments.
${\rm meanRecall} = \frac{1}{n}\sum_{i=1}^{n} {\rm Recall}$. 
It is similar for the other metrics.

\subsection{Implementation Details}
\vspace{.1cm}
\noindent\textbf{Hyper-parameters selection}
The CCTA images are scaled to $v = 0.5$ mm voxel spacing. 
Since the radius of branches usually ranges in $(0,3)$ mm, i.e., $(0,6)$ voxels. 
So the radius of the cube is chosen to be $\gamma=12$ voxels to keep the angel, size, and the texture information. 
Thus the subsampling rate for the centerline positions is $10$ voxels to have overlapping as well as keep the sequential information along the branches. 

\vspace{.1cm}
\noindent\textbf{Training}
The dataset is randomly and equally divided into five subsets. In the training stage, we use a five-fold cross-validation strategy to evaluate all subjects in the dataset. The proposed CPR-GCN model has two trainable components, i.e, 3D CNN following LSTMs (3D CNN module), GCNs following a FC layer (GCN module). A series of 3D image cubes extracted along the vessel centerlines from the 3D CCTA image are the input of the 3D CNN module. The position domain features extracted from the vessel tree via SCT$S^2$ are concatenated with the output of the 3D CNN module. The GCN module takes these combined features as input and predicts the label for every segment. In addition, the reference labels are needed to compute the cross-entropy loss with the predicted labels. The 3D image cubes, position domain features and the reference labels compose the training samples. Our CPR-GCN model is trained in an end-to-end manner. 

The algorithm is implemented using PyTorch with an NVIDIA Tesla P100 GPU. We use Adam optimizer \cite{kingma2014adam} with an initial learning rate of 0.001. Each mini-batch contains 8 coronary artery trees. For each training period, we train the CPR-GCN model up to 200 epochs which takes 2.7 hours. So the total training time for five-fold cross-validation is 13.5 hours. 

\vspace{.1cm}
\noindent\textbf{Testing}
We first choose the best model in each fold according to the overall testing precision ignored classes. Then we use each model to evaluate the corresponding testing data. During inference, the average time spent on the CPR-GCN model is 0.045 s per case, which is greatly important in the clinical utilization. 

\vspace{.1cm}
\noindent\textbf{Results}
Since there is not a public dataset in this field and conventional methods only evaluated their performance on a small private dataset. We also reproduced the conventional method and the deep learning based TreeLab-Net. Considering conventional methods \cite{yang2011automatic,cao2017automatic} mainly rely on registration and prior knowledge, we only reproduce \cite{cao2017automatic} which has a improvement on \cite{yang2011automatic}. Table \ref{table:main_compare} reports the detail performance on our dataset. Our CPR-GCN achieves the highest meanRecall of 0.958, meanPrecision of 0.954 and meanF1 of 0.955, which outperforms other methods with a large margin.

All the models have performed well on the main branches. 
But the side branches are also a crucially important part of the automated anatomical labeling in the CAD system.
In our approach, we treat the main branches and side branches equally. 
So compared with other two-step methods, most of the side branches, such as OM and R-PDA, performs better in our approach.
But since the number of segments of the main branches like LM and RCA is relatively large, the performance of these main branches is better. For the side branches, especially for the D and OM, the number of samples in the dataset is relatively small. So the performance is slightly worse than the main branches.

\subsection{Ablation Study}

\begin{figure}[b]
\setlength{\abovecaptionskip}{-0.1cm}
\setlength{\belowcaptionskip}{-0.5cm}
\centering
\includegraphics[width=0.8\columnwidth]{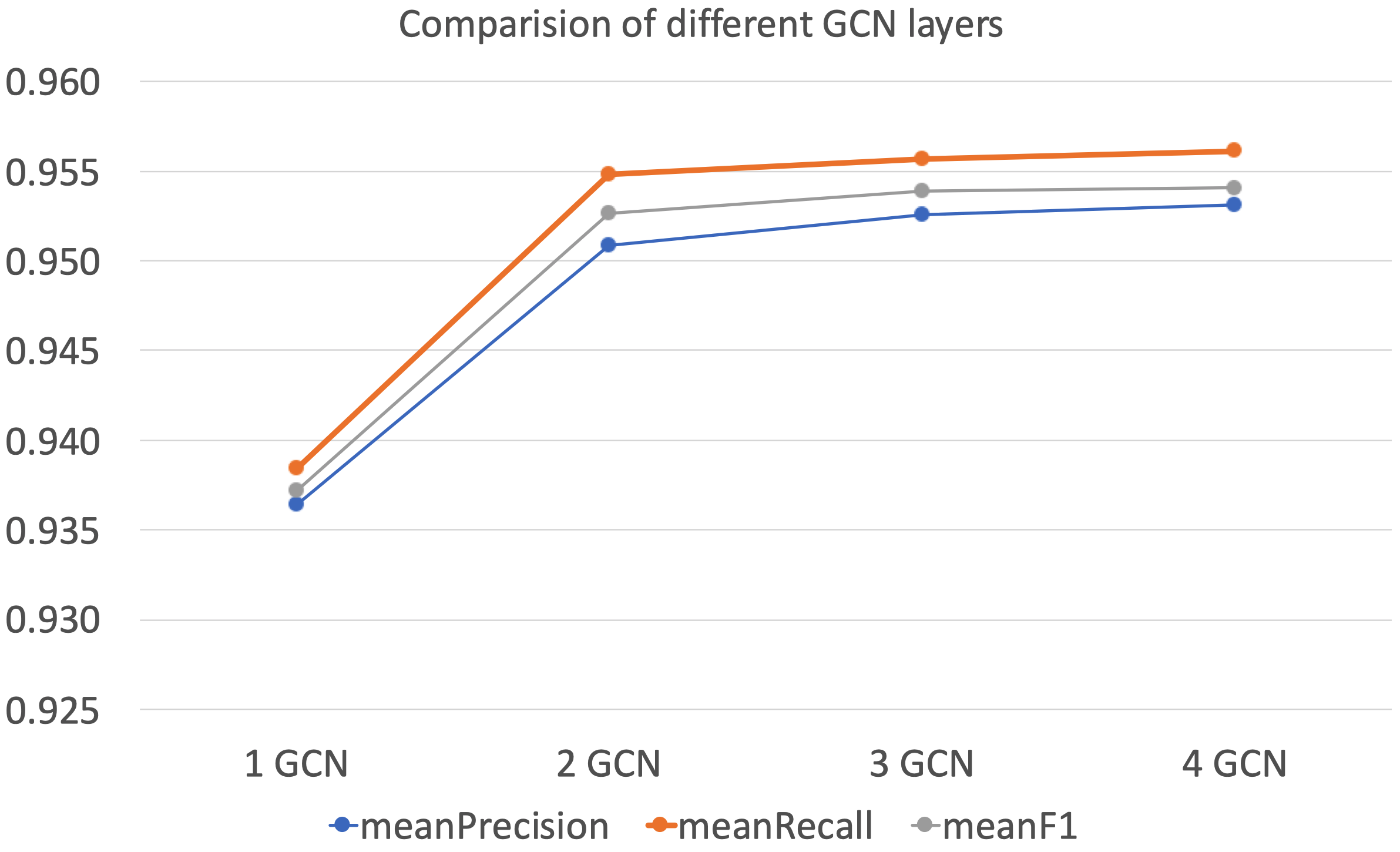}
\caption{Ablation study of repeated GCN blocks. 1 GCN, 2 GCN, 3 GCN and 4 GCN represent stacking corresponding number of the GCN blocks}
\label{fig:result_gcn_layers}
\end{figure}

To make sure that all the components of the CPR-GCN perform well, we design the ablation study experiments.

\vspace{.1cm}
\noindent\textbf{Image domain condition}
One of the major difference between our approach and the other methods is that we use the extra information from CCTA images domain. As shown in the second column of Table \ref{ablationstudy_result}, if we only remove the image domain conditions, the metrics, i.e., meanRecall, meanPrecision and meanF1 will drop over $2.0\%$. For instance, the meanF1 score is $0.934$ compared with $0.955$ in our CPR-GCN.

\vspace{.1cm}
\noindent\textbf{Residual GCN connection}
Also, our approach brought the partial-residual connection in the graph model. In this part, we only remove the residual connection in the GCN block and keep other setting the same as the CPR-GCN. The third column in Table \ref{ablationstudy_result} reports that the F1 score will drop to $0.947$. This argues that with the help from residual, our model can absorb the features from both position and image domain as well as keep the original position domain features.

\vspace{.1cm}
\noindent\textbf{Undirected graph}
In this part, we build the undirected graph, which means adding the opposite edges from the original graph. Detailed results are illustrated in Table \ref{ablationstudy_result}, the fourth column. Although the average metrics among classes (i.e, meanRecall, meanPrecision and meanF1) are slightly worse than our best result, several classes (e.g, LM, LAD, R-PDA, etc.) for undirected graph have higher precision than directed graph. It is worth to note that we select directed graph to achieve higher average metrics. In clinical practice, we can make choice according to the requirement of the doctors.

\vspace{.1cm}
\noindent\textbf{Repeated GCN blocks}
\cite{li2018deeper} reports that stacking multiple GCN layers will result in over-smoothing. We also conduct the experiment to answer how the number of GCN blocks influences the performance. Considering the connection complexity in coronary tree graph, we respectively evaluate our CPR-GCN with 1, 2, 3, 4 GCN layers. Figure \ref{fig:result_gcn_layers} illustrates that we can improve the performance by stacking GCN blocks, especially from 1 GCN block to 2 GCN blocks. However, the model with 4 GCN blocks has no obvious improvement compared with 3 GCN blocks. Therefore we use 3 GCN blocks in our CPR-GCN. The depth of the graph for the coronary artery tree mostly is less than 4, which might lead to the evaluation results. 

\begin{table}[b]
% \footnotesize
\caption{Experimental results of TreeLab-Net \cite{wu2019automated} and our CPR-GCN on the original dataset and synthetic dataset. 20\% RCA and LM is randomly removed in the synthetic dataset.}\smallskip
\centering
\label{table:data_attack}
\resizebox{.99\columnwidth}{!}{
\smallskip\begin{tabular}{c|c|ccc}
\hline
Method & Dataset  & meanRecall &  meanPrecision & meanF1 \\
\hline
\multirow{2}{2.5cm}{TreeLab-Net \cite{wu2019automated}} & Original  & 0.873 & 0.868 & 0.871 \\
& Synthetic  & 0.806 & 0.799 & 0.802 \\
\hline
\multirow{2}{2.5cm}{\textbf{Our CPR-GCN}} & Original  & 0.958 & 0.954 & 0.955 \\
& Synthetic  & 0.931 & 0.928 & 0.929 \\
\hline
\end{tabular}
}
\label{dataset}
\end{table}

\begin{table*}[t]
% \footnotesize
\caption{Part of the ablation study results for our approach. The image domain conditions, as well as the partial-residual connection, are both essential parts of the CPR-GCN.}
\centering
\resizebox{1.0\textwidth}{!}{
\smallskip
\begin{tabular}{ c|ccc|ccc|ccc|ccc}
\hline
\multirow{2}{2.2cm}{Ablation Study} & \multicolumn{3}{c|}{Without image domain conditions} & \multicolumn{3}{c|}{Without residual connection} & \multicolumn{3}{c|}{Undirected graph} & \multicolumn{3}{c}{\textbf{Our CPR-GCN}} \\
& Precision & Recall & F1 Score & Precision & Recall & F1 Score & Precision & Recall & F1 Score & Precision & Recall & F1 Score \\
\hline
LM      & 0.990     & 0.994  & 0.992    & 0.959     & 0.984  & 0.971    & 0.996     & 0.998    & 0.997    & 0.984 & 0.994 & 0.989 \\
LAD     & 0.977     & 0.983  & 0.980    & 0.975     & 0.978  & 0.976    & 0.987     & 0.993    & 0.990    & 0.986 & 0.990 & 0.988 \\
LCX     & 0.938     & 0.964  & 0.951    & 0.946     & 0.963  & 0.955    & 0.963     & 0.980    & 0.971    & 0.971 & 0.982 & 0.977 \\
RI      & 0.880     & 0.904  & 0.892    & 0.917     & 0.871  & 0.893    & 0.875     & 0.904    & 0.890    & 0.896 & 0.921 & 0.909 \\
RCA     & 0.982     & 0.989  & 0.986    & 0.980     & 0.981  & 0.980    & 0.989     & 0.994    & 0.992    & 0.987 & 0.994 & 0.991 \\
D       & 0.842     & 0.877  & 0.859    & 0.883     & 0.934  & 0.908    & 0.889     & 0.924    & 0.906    & 0.883 & 0.936 & 0.909 \\
S       & 0.954     & 0.954  & 0.954    & 0.979     & 0.948  & 0.963    & 0.967     & 0.937    & 0.952    & 0.974 & 0.954 & 0.964 \\
OM      & 0.830     & 0.814  & 0.822    & 0.915     & 0.913  & 0.914    & 0.912     & 0.904    & 0.908    & 0.933 & 0.896 & 0.914 \\
R-PDA   & 0.938     & 0.927  & 0.933    & 0.952     & 0.950  & 0.951    & 0.960     & 0.939    & 0.949    & 0.947 & 0.944 & 0.945 \\
R-PLB   & 0.925     & 0.925  & 0.925    & 0.951     & 0.930  & 0.941    & 0.947     & 0.953    & 0.950    & 0.946 & 0.930 & 0.938 \\
AM      & 0.977     & 0.977  & 0.977    & 0.968     & 0.971  & 0.969    & 0.988     & 0.997    & 0.993    & 0.983 & 0.994 & 0.987 \\
\hline
\bf{Avg.}& 0.930    & 0.937  & 0.934    & 0.948     & 0.947  & 0.947    & 0.952     &  0.957   & 0.954    & \bf 0.954 &\bf 0.958 &\bf 0.955 \\
\bf{Std.}& 0.054    & 0.053  & 0.053    & 0.037     & 0.036  & 0.035    & 0.038     &  0.036   & 0.036    & \bf 0.035 &\bf 0.033 &\bf 0.032 \\
\hline
\end{tabular}
}
\label{ablationstudy_result}
\end{table*}

\subsection{Synthetic "Data Attack"}
We hold the opinion that our CPR-GCN is more robust when main branch in the vessel tree is missing. So we first build a synthetic dataset from our original dataset. 20\% LM and RCA branches is randomly removed. Most of the other side branches (e.g., LCX, LAD, RI, AM, etc.) directly originate from those two branches. In this new synthetic dataset, 295 RCA and LM branches are removed. 1123 of 6760 vessel segments directly connect with these 295 missing branches. Since conventional methods \cite{yang2011automatic,cao2017automatic} strictly rely on the main branches. We only evaluate the trained CPR-GCN and TreeLab-Net \cite{wu2019automated} on this synthetic dataset. As shown in Table \ref{table:data_attack}, our CPR-GCN drops almost 2.6\% while the TreeLab-Net drops almost 6.7\% in three average metrics. This demonstrates that our method is more robust.

%***************************** Discussion *************************************
\section{Discussion}
As shown above, our approach achieves state-of-the-art result. The CPR-GCN takes the image domain information from 3D CCTA images as the conditions for our approach. So as to stress the importance of the features from position domain, we introduced the partial-residual block on the position domain features. 

\vspace{-0.2cm}
\paragraph{Image domain information}
As far as we know, we are the first to include the image domain information as the conditions. 
The result, which is shown above, suggests that even though the position domain features are important components of the automated anatomical labeling of coronary arteries, the CCTA images weigh more than just for semantic segmentation. 
For example, the sizes and the shrinking points are not visible in the centerline-based position features, which may be extracted from the images. 
So even though the positions themselves have relatively rich information in this problem, all the other methods lack the ability to absorb the information from the original 3D CCTA image. 

\vspace{-0.2cm}
\paragraph{Partial-Residual block}
One of the main contributions of our approach is that we brought the partial-residual block. In the traditional residual block, all dimensions of the input $x$ are treated equally. But in the coronary arteries labeling problem, the positions are proven to play an important role. So as to stress this importance as well as use the extra information inside the CCTA images, we use the partial-residual block to treat the image domain in formations as the extra conditions of the model.
With the ablation study, we notice that this kind of structure can improve the metric from $0.947$ to $0.955$. Also, it can make the model more stable.

\vspace{-0.2cm}
\paragraph{Robustness} 
Our CPR-GCN is purely driven by data. So the CPR-GCN can make use of the enormous size of data. Without prior knowledge and the hard-coded ``rule'', the CPR-GCN is more robust to the noises. All the nodes in our graph, which represent different segments, weigh the same in the CPR-GCN. So the wrong classification of LM or other major branches has less chance to spread through all other branches. In order to prove this opinion, we conduct a interesting synthetic "Data Attack" experiment. The results show that missing main branch has less impact on our CPR-GCN than other deep leaning method.

% \vspace{-0.4cm}
\paragraph{Disadvantages and future work}
There are still some future works for this problem. 
Since we treat every branches equally, the imbalance among branches is an issue. It's noticeable that the main branches perform better than side branches due to the number of samples are different. There will also be some tiny branches missing after the segmentation, which will increase the imbalance. Also in this model, we use the discrete approximation of the PDE. In the future, the model can be pushed forward to the continuous model.

%***************************** Conclusion *************************************
\section{Conclusion}
In this paper, we propose the end-to-end conditional partial-residual graph convolutional network (CPR-GCN) model for automated anatomical labeling of coronary arteries, where few alternatives are available. Compared with the traditional methods and the recent deep learning based methods, our approach achieves the state-of-the-art result. We show that with the conditional partial-residual block, both information in position domain and CCTA image domain can be taken into consideration. On the experiment side, we show that our CPR-GCN is more robust and flexible compared with others. The result also shows that the CCTA image domain matters in coronary arteries labeling. Importantly, we show that our algorithmic contributions facilitate the CAD system.

% \newpage

%***************************** References *************************************
{\small
\bibliographystyle{ieee_fullname}
\bibliography{ref}
}

\end{document}